\documentclass[10pt,a4paper]{article}
\usepackage[utf8]{inputenc}
\usepackage[T1]{fontenc}
\usepackage{amsmath}
\usepackage{amsfonts}
\usepackage{amssymb}
\usepackage{graphicx}
\usepackage{natbib}
\usepackage{geometry}
\usepackage{hyperref}
\usepackage{algorithm}
\usepackage{algorithmic}
\usepackage{booktabs}
\usepackage{multirow}
\usepackage{array}
\usepackage{float}

\usepackage{float}
\usepackage{multirow}
\usepackage{threeparttable}
\usepackage{graphicx}
\usepackage{caption}
 \usepackage{subcaption}

\author{
Weiying Zhao$^{1}$, Aleksei Unagaev$^{1}$, Natalia Efremova$^{2}$ \\
$^{1}$Deep Planet \\
$^{2}$Queen Mary University of London\\
}
\date{}

\title{Data-Driven Soil Organic Carbon Sampling: Integrating Spectral Clustering with Conditioned Latin Hypercube Optimization}

\begin{document}

\maketitle

\begin{abstract}

Soil organic carbon (SOC) monitoring often relies on selecting representative field sampling locations based on environmental covariates. We propose a novel hybrid methodology that integrates {spectral clustering}—an unsupervised machine learning technique—with {conditioned Latin hypercube sampling} (cLHS) to enhance the representativeness of SOC sampling. In our approach, spectral clustering partitions the study area into $K$ homogeneous zones using multivariate covariate data, and cLHS is then applied within each zone to select sampling locations that collectively capture the full diversity of environmental conditions. This hybrid spectral-cLHS method ensures that even minor but important environmental clusters are sampled, addressing a key limitation of vanilla cLHS which can overlook such areas. We demonstrate on a real SOC mapping dataset that spectral-cLHS provides more uniform coverage of covariate feature space and spatial heterogeneity than standard cLHS. This improved sampling design has the potential to yield more accurate SOC predictions by providing better-balanced training data for machine learning models.
\end{abstract}

\section{Introduction}
Monitoring and mapping {soil organic carbon (SOC)} is crucial for understanding soil health, climate change mitigation, and sustainable land management \cite{alison2019environment}. However, obtaining SOC measurements is resource-intensive, so it is essential to carefully select a limited number of field sampling locations that adequately represent the variability of environmental conditions across a landscape. Traditional sampling designs (e.g., random or grid sampling) may either miss important variability or be inefficient in capturing complex multivariate patterns in the environment \cite{campbell2022review}. Efficient agricultural management requires smart sampling schemes that maximize information gain from each sample.

One such scheme is {conditioned Latin hypercube sampling (cLHS)}, introduced in \cite{minasny2006conditioned}. The cLHS algorithm selects sample locations such that the distributions of environmental covariates (e.g., spectral indices, soil maps, elevation) in the sample set closely resemble those in the entire area of interest. By stratifying sampling across multiple covariates simultaneously, cLHS produces a set of points that are efficiently distributed in the environmental feature space, improving the chances that all relevant conditions for SOC are observed. cLHS has been widely applied in digital soil mapping and environmental monitoring studies due to its ability to capture multivariate variability efficiently.

Despite its strengths, vanilla cLHS has limitations. Because it optimizes sample selection globally, it may overlook smaller regions of the landscape that have unusual combinations of covariate values (i.e., minority classes or outliers in feature space). These areas might be ecologically or agronomically important (for instance, pockets of high SOC or unique soil types) but could remain unsampled if they occupy only a small fraction of the study area. Additionally, cLHS does not explicitly account for spatial structure beyond what is encoded in the covariates, potentially leading to spatial clustering of sample points in areas with high covariate density and leaving other areas sparse.

To address these challenges, we propose a hybrid approach that integrates unsupervised clustering with cLHS. In particular, we employ {spectral clustering} on the environmental covariates to partition the study area into $K$ relatively homogeneous zones prior to sampling. Spectral clustering is a graph-based clustering technique effective at capturing complex, non-linear relationships in data by embedding points in a lower-dimensional space using eigenvectors of a similarity matrix \cite{ng2001spectral}. By segmenting the landscape into meaningful clusters (or zones), we can ensure that subsequent sampling allocates attention to all distinct regions of feature space. We then perform cLHS within each cluster (zone) rather than on the whole area at once. This guarantees that each cluster contributes at least one sample (or a proportional number of samples) to the final set, thereby covering even the previously under-represented parts of feature space. Our approach, which we refer to as {spectral-cLHS}, effectively combines the strengths of clustering (to delineate diverse strata) and cLHS (to sample optimally within each stratum).

In this paper, we present the spectral-cLHS methodology and evaluate its performance against standard cLHS for SOC monitoring. We first describe the study area and data, as well as the implementation of spectral clustering and cLHS in our workflow (Section \ref{sec:methods}). We then compare the sampling results of spectral-cLHS and vanilla cLHS, demonstrating how the integrated approach yields a more comprehensive coverage of environmental variation (Section \ref{sec:results}). Finally, we discuss the implications of this hybrid sampling strategy for SOC prediction and broader environmental data collection efforts, and conclude with potential directions for future work (Section \ref{sec:conclusion}).

\section{Methodology}\label{sec:methods}
\subsection{Study Area and Covariate Data}
The proposed approach was tested on a SOC mapping study area comprising a heterogeneous agricultural landscape. The study region is characterized by diverse soil types, considerable topographic variation, and heterogeneous vegetation cover, reflecting a wide range of conditions influencing SOC. A set of environmental covariate layers \cite{zhao2023soil, zhao2024soil} was assembled to represent factors influencing SOC distribution, including remote sensing data, soil and terrain data. We employ multi-temporal Sentinel-2 satellite imagery (multiple dates) and a recent single-date Sentinel-2 image, processed to surface reflectance. These provide information on vegetation indices, moisture, and other land surface properties. Additionally, we use a digital elevation model (DEM) and derivatives (slope, aspect) capturing topographic effects, and soil property maps (soil texture and existing SOC estimates from SoilGrids and similar datasets).

\begin{figure}[H]
  \centering
  \includegraphics[width=0.9\textwidth]{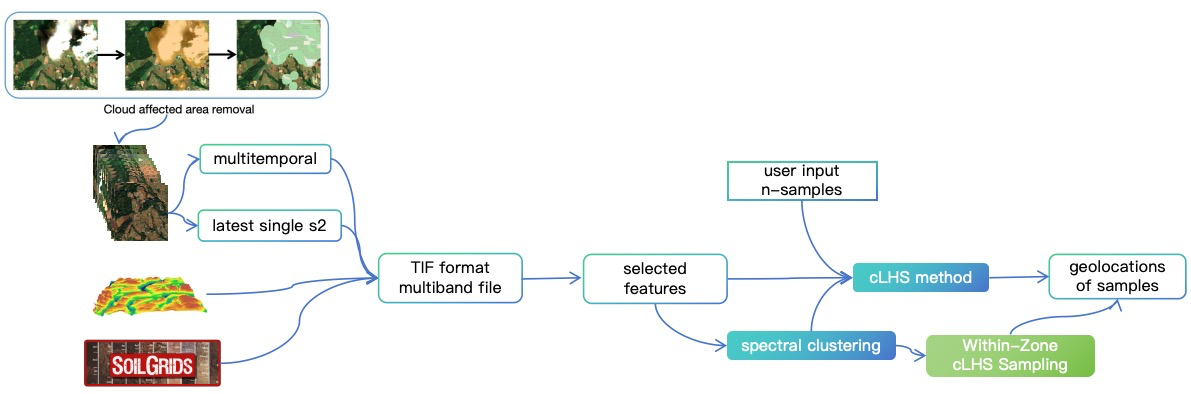}
  \caption{Sampling location selection using the Conditioned Latin Hypercube Sampling method (cLHS). The number of samples can be input or computed using the spectral clustering method based on the selected features. cLHS method can process independently or within each zone generated by spectral clustering.}
  \label{fig:flowchart}
\end{figure}

All covariate layers were resampled to a common grid and stacked into a multivariate feature dataset (in GeoTIFF format). Each grid cell in the study area was represented by a feature vector (e.g., reflectances, indices, elevation, etc.). Prior to clustering and sampling, we applied standard preprocessing: cloud-affected pixels in the satellite data were masked out or removed, and features were normalized to comparable scales. The overall workflow of our methodology is illustrated in Figure \ref{fig:flowchart}.


\subsection{Spectral Clustering for Zone Delineation}
We use spectral clustering to divide the study area into $K$ clusters based on the covariate feature vectors. In spectral clustering, data points are treated as nodes in a graph, with edge weights defining similarity (for example, an RBF kernel on the feature distance). We construct a similarity matrix $W$ where $W_{ij}$ reflects the proximity of feature vectors for grid cells $i$ and $j$. To make the problem computationally tractable for a large spatial dataset, we employed a sampled approach: computing $W$ for a subset of points or using approximate neighbors.

\begin{figure}[tb]
\centering
\includegraphics[width=0.9\textwidth]{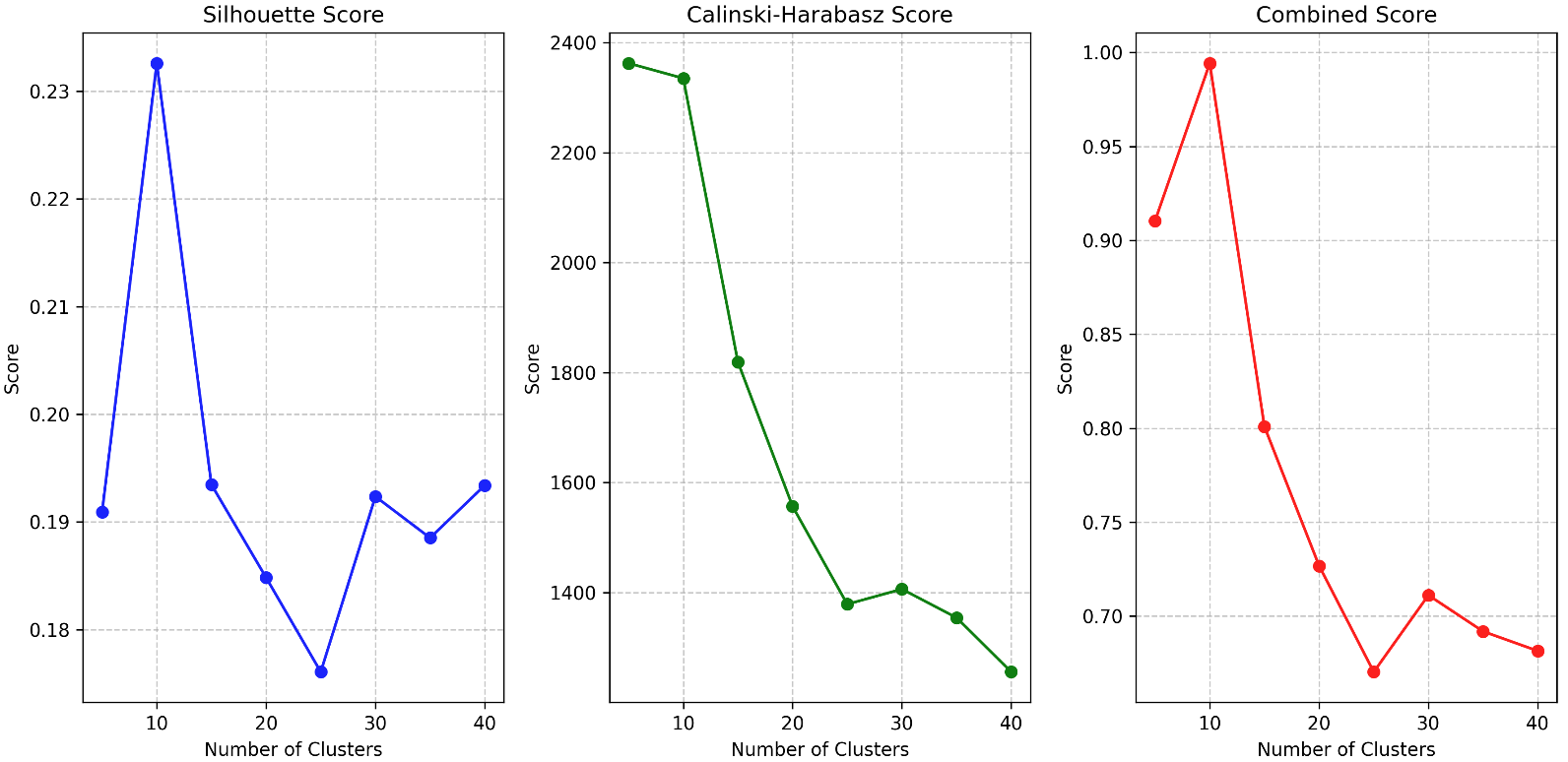}
\caption{Cluster validity metrics as a function of number of clusters $K$. \textit{Left:} Mean Silhouette coefficient (blue curve), which peaks at $K=10$. \textit{Center:} Calinski-Harabasz index (green curve), which is also relatively high at $K=10$ (indicating well-separated clusters). \textit{Right:} Combined normalized score (red) used to select the optimal $K$. In this case, $K=10$ was chosen as the optimal number of clusters.}
\label{fig:cluster_eval}
\end{figure}

Following the method of \cite{ng2001spectral}, we compute the normalized graph Laplacian $L$ and its $K$ leading eigenvectors. These eigenvectors, when used as $K$-dimensional embeddings of the data, are clustered (e.g., by $k$-means) to yield the final partition of the $N$ grid cells into $K$ clusters. Each cluster represents a zone of relatively homogeneous covariate characteristics. An important decision in this process is choosing the number of clusters $K$. We evaluated cluster validity indices including the \textbf{Silhouette coefficient} and the \textbf{Calinski-Harabasz (CH) index} for different values of $K$. Figure \ref{fig:cluster_eval} shows an example of these metrics, where the Silhouette score (higher is better) and CH score (higher is better) suggested an optimal clustering around $K=10$ for our dataset. We combined the two indices (after normalization) into a single composite score to confirm that $K=10$ maximized overall clustering quality (Figure \ref{fig:cluster_eval}, right panel).

\begin{figure}[tb]
\centering
\includegraphics[width=0.85\textwidth]{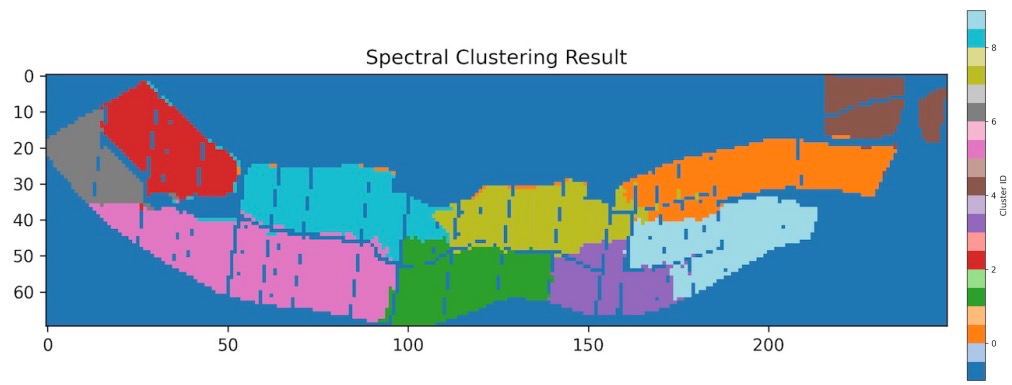}
\caption{Result of spectral clustering on the study area covariates, yielding $K=10$ clusters (zones). Each color corresponds to a different cluster (cluster IDs 0--9). This clustered map highlights distinct regions of the landscape with homogeneous feature characteristics.}
\label{fig:clusters_map}
\end{figure}

\begin{figure}[tb]
    \centering
    \begin{subfigure}[b]{0.45\textwidth}
        \centering
        \includegraphics[width=\textwidth]{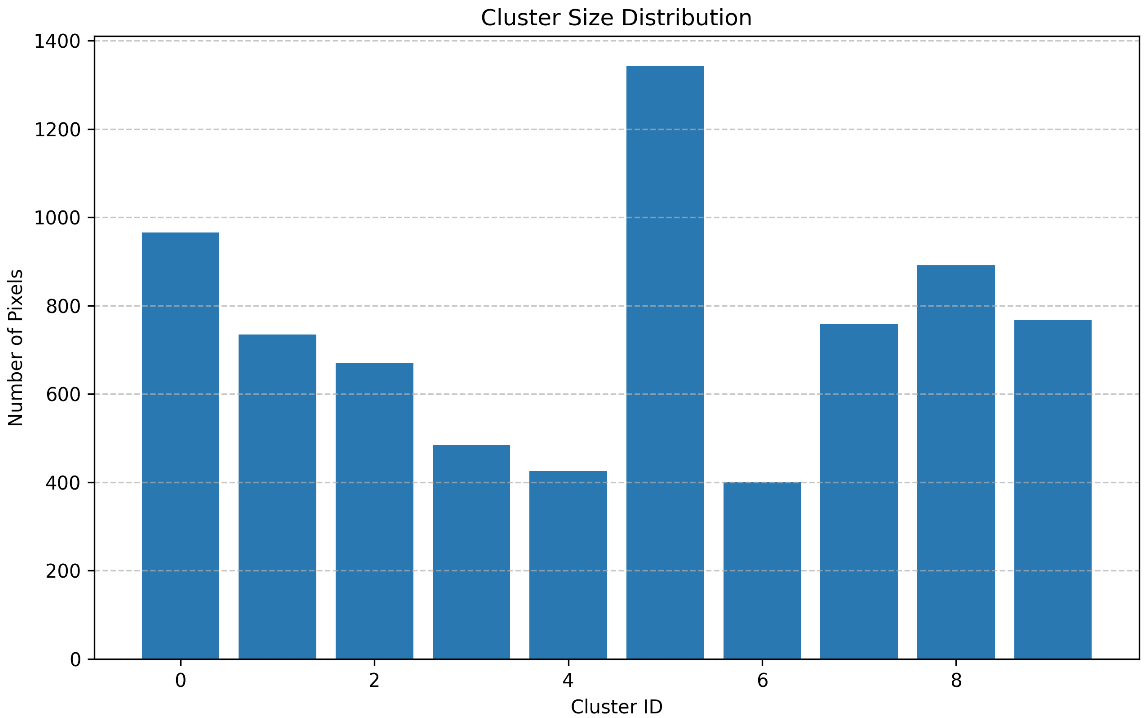}
        \caption{Pixel count per spectral cluster (k = 10)}
        \label{fig:cluster_size}
    \end{subfigure}
    \hfill
    \begin{subfigure}[b]{0.45\textwidth}
        \centering
        \includegraphics[width=\textwidth]{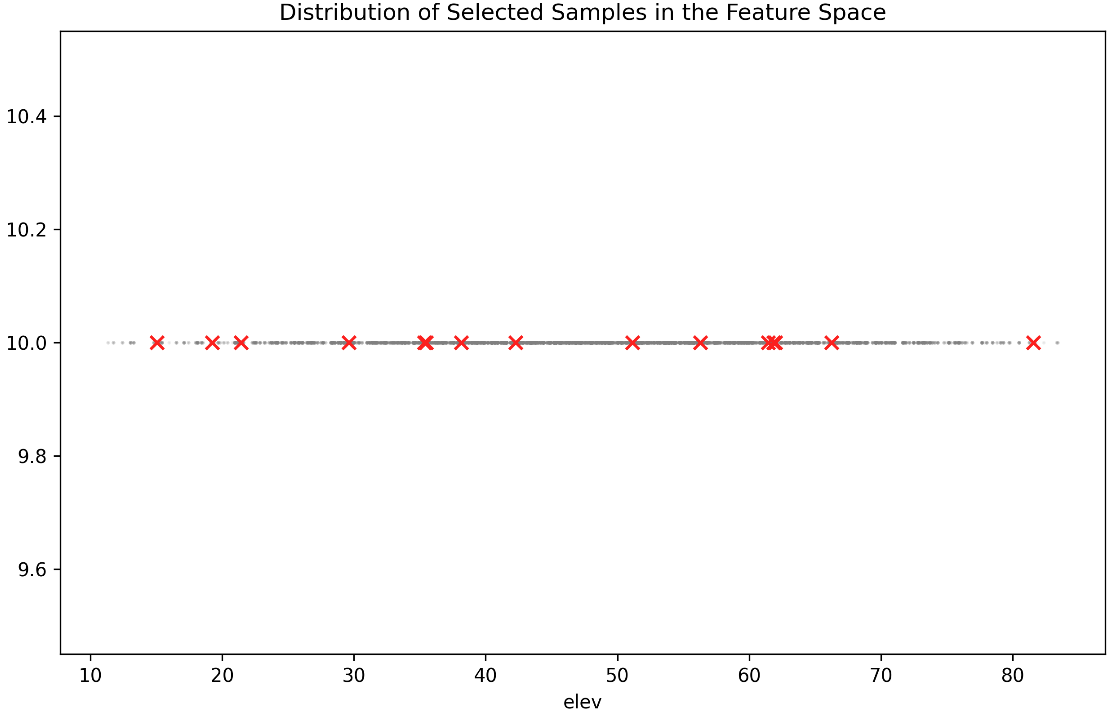}
        \caption{Sample coverage in simple feature space. }
        \label{fig:sample_dem_locations}
    \end{subfigure}
    \caption{Cluster feature analysis. \textbf{(a) }
Cluster size distribution for $K=10$ clusters, showing the number of grid cells in each cluster (Cluster ID 0 through 9). The cluster sizes vary, with some clusters covering a much larger area (e.g., cluster 5) and others representing relatively small zones (e.g., cluster 6). This justifies allocating at least one sample to each cluster to ensure even small clusters are represented in the sampling. \textbf{(b)}
Grey points represent all environmentally valid pixels plotted as elevation distribution;
red crosses mark the cLHS-selected sampling locations.
The samples cover the full 10–80\,m elevation gradient without clumping, demonstrating that
the final design is representative along this key terrain axis.}
    \label{fig:cluster_features}
\end{figure}

Applying spectral clustering with $K=10$ to the full dataset yielded a map of cluster zones (Figure \ref{fig:clusters_map}). Each cluster zone is a contiguous or discrete set of locations that share similar covariate patterns. The clusters vary in spatial extent and environmental characteristics. For instance, some clusters correspond to low-lying areas with high moisture and potentially higher SOC, while others identify drier upland zones with different spectral signatures. Figure \ref{fig:cluster_size} summarizes the size (number of grid cells) of each cluster. We observe that the largest cluster (ID 5) contains over 1300 cells, whereas the smallest cluster (e.g., ID 6) contains under 500 cells. This indicates the presence of both widespread environmental conditions and more localized, rare ones in the study area. By using clustering, our method explicitly recognizes these smaller zones that might be missed by a purely global sampling approach.

\subsection{Within-Zone cLHS Sampling}
After delineating the clusters, we integrate this clustering with cLHS. The user decides on a total number of samples $n$ to collect based on resources. Instead of applying cLHS on the entire study area for $n$ samples, we divide the $n$ across the $K$ clusters. In our implementation, we allocated samples to clusters roughly proportional to their size (number of grid cells), while ensuring every cluster received at least one sample. For example, if $n=100$ and cluster $i$ contains 10\% of the area’s grid cells, cluster $i$ would receive about 10 samples (rounded to the nearest integer).

Within each cluster, we then run the standard cLHS algorithm to select specific grid cells for sampling. Each cluster’s cLHS is constrained to only that cluster’s subset of data points and aims to make those local samples representative of that cluster’s internal covariate distributions. We utilized an existing implementation of cLHS to perform this optimization. The cLHS algorithm tries to maximize the spread of samples in the multivariate space of covariates: it typically uses a heuristic (such as simulated annealing) to minimize a cost function that measures the difference between the distribution of covariate values in the selected samples and the distribution in the entire population (here, the cluster). By applying this within each cluster, we ensure that for every environmental zone identified, the chosen samples mirror the range of conditions present in that zone.

Finally, the sample sets from all clusters are combined to form the full sampling design. This final set of $n$ sample locations is then proposed for field sampling and SOC measurement. The hybrid approach guarantees representation of each cluster while still preserving the core idea of cLHS to approximate covariate distributions. We refer to this multi-step procedure as \textbf{spectral-cLHS sampling}. In contrast, a \textbf{vanilla cLHS} approach would simply ignore cluster boundaries and pick $n$ samples from the whole area in one optimization run (Fig. \ref{fig:sample_maps}.

\section{Results and Discussion}\label{sec:results}
We compare the spectral-cLHS approach to vanilla cLHS by examining the spatial distribution of selected samples and their coverage of the environmental feature space. The total sample size in our demonstration was set to $n=10$ for ease of visualization on maps (one sample per cluster in the spectral-cLHS case), but the method is equally applicable to larger $n$.

\begin{figure}[tb]
\centering
\begin{subfigure}{0.8\textwidth}
\centering
\includegraphics[width=\textwidth]{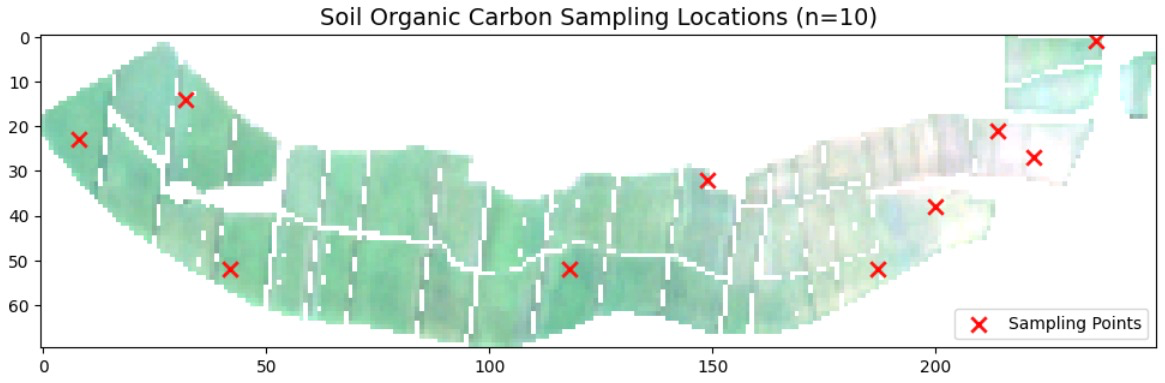}
\caption{Standard cLHS (global) sampling.}
\end{subfigure}
\hfill
\begin{subfigure}{0.8\textwidth}
\centering
\includegraphics[width=\textwidth]{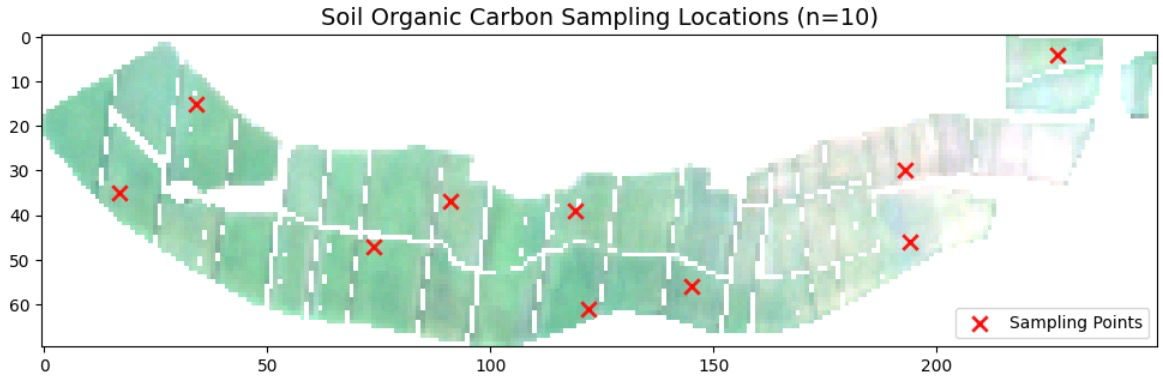}
\caption{Proposed spectral-cLHS sampling.}
\end{subfigure}
\caption{Comparison of sampling locations (red \textbf{x} markers) selected by (a) vanilla cLHS and (b) the hybrid spectral-cLHS method for a total of $n=10$ samples. The background shows the spectral clustering zones (colored regions, same as Figure \ref{fig:clusters_map}). Standard cLHS places multiple samples in some large clusters while missing some smaller clusters entirely, whereas spectral-cLHS guarantees at least one sample per cluster, achieving a more balanced spatial coverage.}
\label{fig:sample_maps}
\end{figure}

\paragraph{Spatial distribution of samples:} in Figure \ref{fig:sample_maps}, we compare the sample locations obtained from the two methods. In the traditional cLHS approach (Figure \ref{fig:sample_maps}a), the 10 selected points (red \textbf{x} markers) tend to fall in a few regions of the map, leaving some clusters  with no samples. This outcome is due to the global cLHS algorithm favoring areas that dominate the overall covariate distribution, which can inadvertently ignore smaller clusters. In contrast, the spectral-cLHS design (Figure \ref{fig:sample_maps}b) includes exactly one sample in each of the 10 clusters. The red \textbf{x} markers are more evenly dispersed, and every colored cluster region contains at least one sample. This ensures that even the previously under-sampled zones (such as cluster 6 in the center and cluster 8 on the right in our example) are now represented in the sample set.

The improvement in spatial coverage is evident: spectral-cLHS effectively acts as a stratified sampling scheme where clusters are strata, thereby combining the strengths of stratification with the multi-variable balancing of cLHS. This design can be especially advantageous when certain clusters correspond to areas of particular interest (e.g., zones expected to have extreme SOC values or unique land management practices) because it ensures those zones are not overlooked.

\begin{figure}[tb]
\centering
\includegraphics[width=0.75\textwidth]{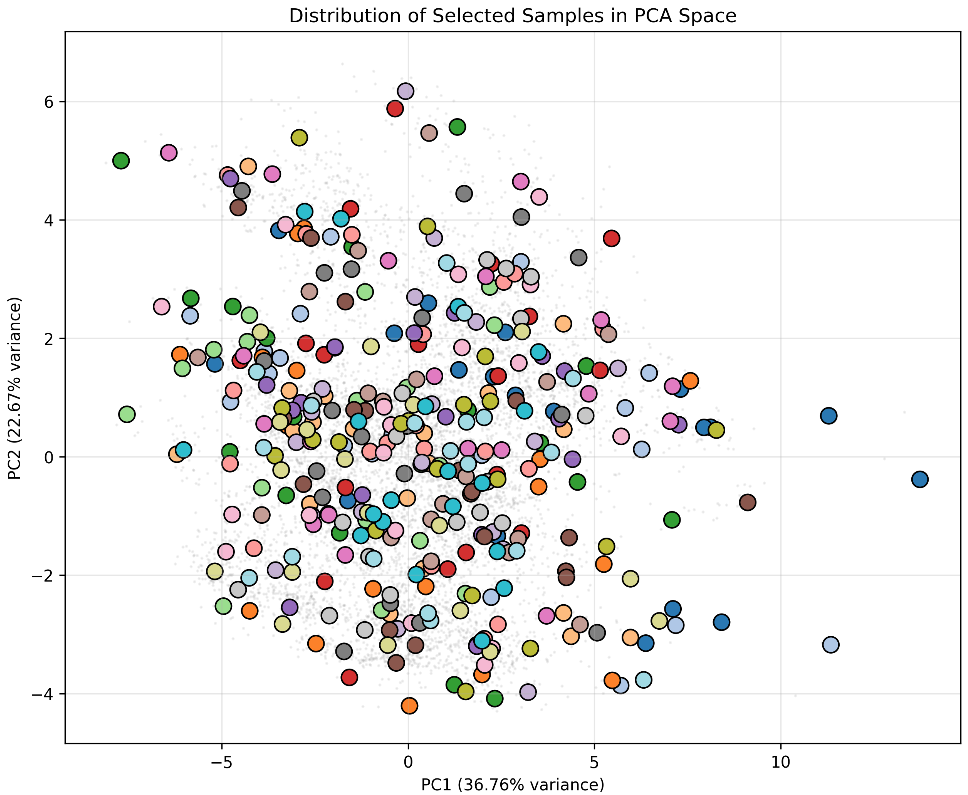}
\caption{Distribution of selected samples in the space of the first two principal components (PC1 and PC2) of the covariate feature set. Grey dots show all data points (grid cells) in PCA space, and the colored circles show the samples chosen by spectral-cLHS (each sample colored by its cluster ID). The samples from spectral-cLHS span the range of the data cloud, indicating that all major directions of variance in the feature space are represented. By ensuring at least one sample per cluster, the method captures both the high-density central regions and the more peripheral regions of the feature space, which a vanilla cLHS might under-sample.}
\label{fig:pca_space}
\end{figure}

\paragraph{Coverage of feature space:} We also assessed how well each sampling strategy covers the multivariate feature space of the covariates. Figure \ref{fig:pca_space} illustrates the distribution of selected samples in the space of the first two principal components (PC1 and PC2) of the covariate data. The background grey points represent all grid cells projected into this PCA space, showing the continuous distribution of environmental conditions. The colored circles indicate the chosen samples by the spectral-cLHS method, with each sample colored according to its cluster membership (matching the cluster colors in Figure \ref{fig:clusters_map}). We see that the spectral-cLHS samples are spread across the range of PC1 and PC2, effectively covering the broad cloud of data points. Each cluster contributes at least one sample, so even outlying portions of the feature space  have a representative sample. In contrast, a single-run cLHS (not shown in the figure) might select samples that cluster in the densest part of feature space, potentially missing some extremes or less frequent combinations of covariate values.

The broader coverage of feature space by spectral-cLHS is expected to improve model performance in SOC prediction. With samples spanning all clusters, a predictive model (e.g., a random forest or neural network) trained on these samples would have exposure to all types of environments present in the area, potentially leading to more robust predictions. In contrast, if some environmental conditions are missing from the training data (as could happen with vanilla cLHS), the model might perform poorly in those unsampled conditions. Although an evaluation of prediction accuracy is beyond the scope of this workshop paper, the sampling design improvements we demonstrate are a crucial first step toward more reliable SOC mapping.

Another advantage of the spectral-cLHS approach is its flexibility. If certain clusters are deemed more important (for example, due to stakeholder interest or known higher SOC variance), the user can allocate more samples to those clusters while still sampling all clusters. This method also inherently parallelizes the sampling problem: since cLHS is executed independently within each cluster, the computational problem is broken into smaller pieces. This can be more efficient than a single large cLHS run, and it allows different clusters’ sampling to be performed in parallel or with different parameter settings if needed.

\section{Conclusion}\label{sec:conclusion}
We have presented a novel hybrid sampling methodology, spectral-cLHS, that integrates spectral clustering with cLHS to optimise soil organic carbon monitoring. By first partitioning the landscape into homogeneous clusters, then allocating and selecting samples within each cluster, our approach ensures comprehensive coverage of both spatial and feature-space heterogeneity. In an application to a diverse agricultural landscape, spectral-cLHS proved effective in capturing minority environmental conditions that a traditional cLHS might miss, resulting in a more balanced and representative set of sampling locations. The implications of this method are significant for machine learning applications in environmental science. Providing a learning algorithm with training data that cover all relevant conditions will enhance the generalization properties of the spectral-cLHS method. The proposed approach can improve field sampling, and the accuracy of SOC predictive models and maps, which are vital for carbon accounting and land management decisions.

Future work will involve applying the spectral-cLHS approach in different geographic settings and for different target variables (e.g., other soil properties or vegetation indices) to further validate its generalisation properties. We also plan to explore automated ways of determining the optimal number of clusters $K$ and integrating expert knowledge (such as known soil classifications) into the clustering step. Finally, a direct comparison of model prediction performance using training data from spectral-cLHS vs. traditional designs would quantify the practical benefits of the proposed method. We believe that hybrid approaches like spectral-cLHS, which blend unsupervised learning and smart sampling, can greatly enhance data efficiency in environmental monitoring and beyond.

\nocite{*}
\bibliography{bib_conference.bib}

\end{document}